\documentclass{article}
\usepackage{ijcai16}
\usepackage{helvet}
\usepackage{courier}
\usepackage{amssymb}
\usepackage{amsmath}
\usepackage{graphicx}
\usepackage{multirow}
\usepackage{floatrow}
\usepackage{times}

\title{Unsupervised Word and Dependency Path Embeddings \center{for Aspect Term Extraction}}
\author{Yichun Yin$^{1}$, Furu Wei$^{2}$, Li Dong$^{3}$, Kaimeng Xu$^{1}$, Ming Zhang$^{1}$\thanks{Corresponding author: Ming Zhang}, Ming Zhou$^{2}$ \\ 
	$^{1}$School of EECS, Peking University  \\
	$^{2}$Microsoft Research \\
	$^{3}$Institute for Language, Cognition and Computation, University of Edinburgh \\
	\{yichunyin,1300012834,mzhang\_cs\}@pku.edu.cn,\{fuwei,mingzhou\}@microsoft.com,li.dong@ed.ac.uk}
	
\begin{document}

		\newcommand{\namecite}[1]{\citeauthor{#1}~\shortcite{#1}}
		\newcommand{\onenamecite}[2]{\citeauthor{#1}~\shortcite{#1,#2}}
		\maketitle
		\let\oldsqsubset\sqsubset
		\renewcommand{\sqsubset}[1][0pt]{%
			\mathrel{\raisebox{#1}{$\oldsqsubset$}}%
		}
		\let\olddagger\dagger
		\renewcommand{\dagger}[1][0pt]{%
			\mathrel{\scalebox{0.75}{\raisebox{#1}{$\olddagger$}}}%
		}
		\let\oldast\ast
		\renewcommand{\ast}[1][0pt]{%
			\mathrel{\scalebox{0.75}{\raisebox{#1}{$\oldast$}}}%
		}	
			
		\begin{abstract}
			In this paper, we develop a novel approach to aspect term extraction based on unsupervised learning of distributed representations of words and dependency paths. The basic idea is to connect two words ($w_1$ and $w_2$) with the dependency path ($r$) between them in the embedding space. Specifically, our method optimizes the objective $\textbf{w}_1 + \textbf{r} \approx \textbf{w}_2$ in the low-dimensional space, where the multi-hop dependency paths are treated as a sequence of grammatical relations and modeled by a recurrent neural network. Then, we design the embedding features that consider linear context and dependency context information, for the conditional random field (CRF) based aspect term extraction. Experimental results on the SemEval datasets show that, (1) with only embedding features, we can achieve state-of-the-art results; (2) our embedding method which incorporates the syntactic information among words yields better performance than other representative ones in aspect term extraction.
		\end{abstract}

		\section{Introduction}
		
		Aspect term extraction~\cite{hu2004mining,pontiki2014semeval,pontiki2015semeval} aims to identify the aspect expressions which refer to the product's or service's properties (or attributes), from the review sentence.
		It is a fundamental step to obtain the fine-grained sentiment of specific aspects of a product, besides the coarse-grained overall sentiment. Until now, there have been two major approaches for aspect term extraction. The unsupervised (or rule based) methods \cite{qiu2011opinion} rely on a set of manually defined opinion words as seeds and rules derived from syntactic parsing trees to iteratively extract aspect terms. The supervised methods \cite{jakob2010extracting,li2010structure,chernyshevich2014ihs,zhiqiang2014dlirec,sanvicente-saralegi-agerri:2015:SemEval} usually treat aspect term extraction as a sequence labeling problem, and conditional random field (CRF) has been the mainstream method in the aspect term extraction task of SemEval.
		
		Representation learning has been introduced and achieved success in natural language processing (NLP) \cite{6472238}, such as word embeddings \cite{mikolov2013distributed} and structured embeddings of knowledge bases~\cite{bordes2011learning}. It learns distributed representations for text in different granularities, such as words, phrases and sentences, and reduces data sparsity compared with the conventional one-hot representation. The distributed representations have been reported to be useful in many NLP tasks \cite{turian2010word,collobert:2011b}.
		
		In this paper, we focus on representation learning for aspect term extraction under an unsupervised framework. Besides words, dependency paths, which have been shown to be important clues in aspect term extraction~\cite{qiu2011opinion}, are also taken into consideration. Inspired by the representation learning of knowledge bases \cite{bordes2011learning,neelakantan-roth-mccallum:2015:ACL-IJCNLP,Lin2015Model} that embeds both entities and relations into a low-dimensional space, we learn distributed representations of words and dependency paths from the text corpus. Specifically, the optimization objective is formalized as $\textbf{w}_1 + \textbf{r} \approx \textbf{w}_2$. In the triple $(w_1,w_2,r)$, $w_1$ and $w_2$ are words, $r$ is the corresponding dependency path consisting of a sequence of grammatical relations.
		The recurrent neural network \cite{mikolov2010recurrent} is used to learn the distributed representations of dependency paths. Furthermore, the word embeddings are enhanced by linear context information in a multi-task learning manner.

		The learned embeddings of words and dependency paths are utilized as features in CRF for aspect term extraction. The embeddings are real values that are not necessarily in a bounded range \cite{turian2010word}.
		We therefore firstly map the continuous embeddings into the discrete embeddings and make them more appropriate for the CRF model. 
		Then, we construct the embedding features which include the target word embedding, linear context embedding and dependency context embedding for aspect term extraction. We conduct experiments on the SemEval datasets and obtain comparable performances with the top systems. To demonstrate the effectiveness of the proposed embedding method, we also compare our method with other state-of-the-art models. With the same feature settings, our approach achieves better results. Moreover, we perform a qualitative analysis to show the effectiveness of the learned word and dependency path embeddings.
		
		The contributions of this paper are two-fold. 
		First, we use the dependency path to link words in the embedding space for distributed representation learning of words and dependency paths. By this method, the syntactic information is encoded in word embeddings and the multi-hop dependency path embeddings are learned explicitly. Second, we construct the CRF features only based on the derived embeddings for aspect term extraction, and achieve state-of-the-art results.
	
		\section{Related Work}
		There are a number of unsupervised methods for aspect term extraction. \namecite{hu2004mining} mine aspect term based on a data mining approach called association rule mining. In addition, they use opinion words to extract infrequent aspect terms. Using relationships between opinion words and aspect words to extract aspect term is employed in many follow-up studies. In \cite{qiu2011opinion}, the dependency relation is used as a crucial clue, and the double propagation method is proposed to iteratively extract aspect terms and opinion words.
		
		The supervised algorithms \cite{li2012cross,liu-joty-meng:2015:EMNLP} have been provided for aspect term extraction. Among these algorithms, the mainstream method is the CRF model. \namecite{li2010structure} propose a new machine learning framework on top of CRF to jointly extract positive opinion words, negative opinion words and aspect terms. In order to obtain a more domain independent extraction, \namecite{jakob2010extracting} train the CRF model on review sentences from different domains. Most of top systems in SemEval also rely on the CRF model, and hand-crafted features are constructed to boost performances.
		
		The representation learning has been employed in NLP~\cite{6472238}. Two successful applications are the word embedding~\cite{mikolov2013distributed} and the knowledge embedding~\cite{bordes2011learning}. The word embeddings are learned from raw textual data and are effective representations for word meanings. They have been proven useful in many NLP tasks~\cite{turian2010word,guo2014revisiting}. The knowledge embedding method models the relation as a translation vector that connects the vectors of two entities. It optimizes an objective over all the facts in the knowledge bases, encoding the global information. The learned embeddings are useful for reasoning missing facts and facts extraction in knowledge bases.
		
		As dependency paths contain rich linguistics information between words, recent works model it as a dense vector. \namecite{liu-EtAl:2015:ACL-IJCNLP} incorporate the vectors of grammatical relations (one-hop dependency paths) into the relation representation between two entities for relation classification task. 
		Dependency-based word embedding model \cite{levy-goldberg:2014:P14-2} encodes dependency information into word embeddings which is one similar work to ours. However, it implicitly encodes the dependency information, embedding the unit \emph{word + dependency path} as the context vector. Besides, it only considers one-hop dependency paths and ignores multi-hop dependency paths. In this paper, we take the Dependency-based word embedding as a baseline method.
		
		The approaches that learn path embeddings are investigated in knowledge bases. 
		\namecite{neelakantan-roth-mccallum:2015:ACL-IJCNLP} provide a method that reasons about conjunctions of multi-hop relations for knowledge base completion. The implication of a path is composed by using a recurrent neural network. \namecite{Lin2015Model} also learn the representations of relation paths. A path-constraint resource allocation algorithm is employed to measure the reliability of path. In this paper, we learn the semantic composition of dependency paths over dependency trees.		
			\begin{figure*}
						\centering
						\includegraphics[width=0.80\textwidth]{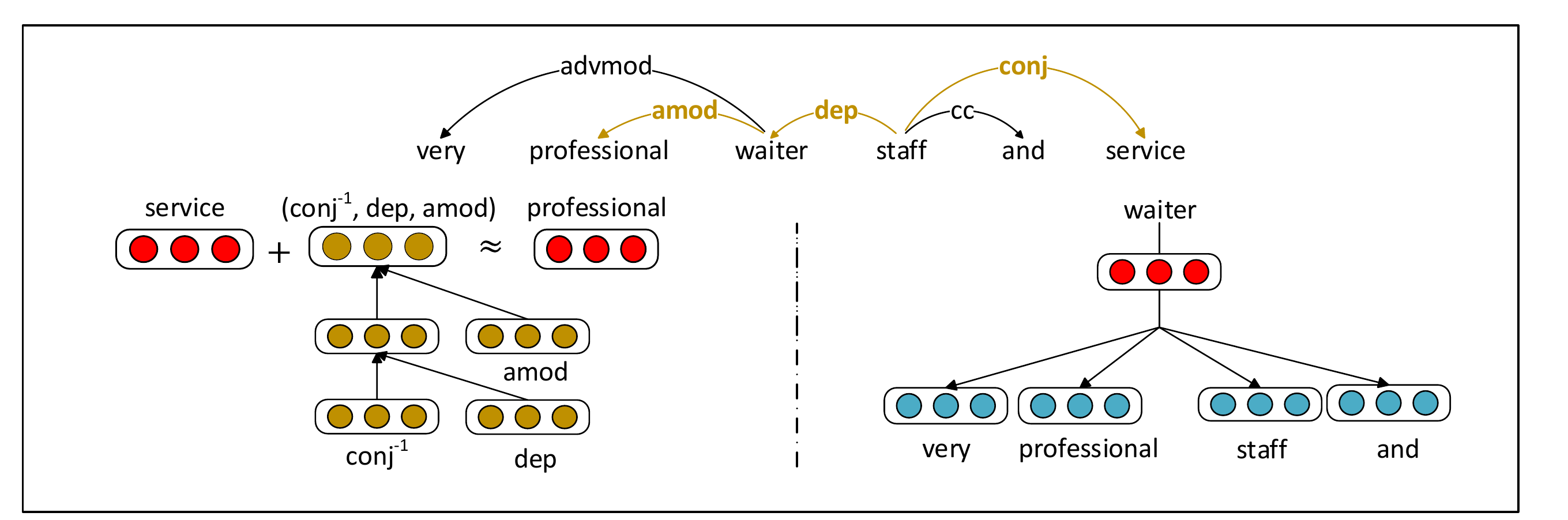}
						\caption{The overview of our unsupervised learning method. Left: Unsupervised learning of word and dependency path embeddings. Right: Multi-task learning with linear context.}
						\label{fig:Figure1}
			\end{figure*}
		
		\section{Method}		
		In this section, we first present the unsupervised learning of word and dependency path embeddings. In addition, we describe how to enhance word embeddings by leveraging the linear context in a multi-task learning manner. Then, the construction of embedding features for CRF model is discussed.
		
		\subsection{Unsupervised Learning of Word and Dependency Path Embeddings}
		We learn the distributed representations of words and dependency paths in a unified model, as shown in Figure 1. We extract the triple $(w_1,w_2,r)$ from dependency trees, where $w_1$ and $w_2$ denote two words, the corresponding dependency path $r$ is the shortest path from $w_1$ to $w_2$ and consists of a sequence of grammatical relations. For example, in Figure 1, given that \emph{service} is $w_1$ and \emph{professional} is $w_2$, we obtain the dependency path $ *\stackrel{conj}{\longleftarrow} *  \stackrel{dep}{\longrightarrow} *  \stackrel{amod}{\longrightarrow} * $. We notice that considering the lexicalized dependency paths can provide more information for the embedding learning. However, we need to memorize more dependency path frequencies for the learning method (negative sampling) described in Section 3.3. Formally, the number of dependency paths is $|V_{word}|^{n-1}|V_{dep}|^{n}$ when we take n-hop dependency paths into consideration\footnote{$V_{word}$ is the set of words and $|V_{word}|$ is over 100,000 in our paper. $V_{dep}$ is the set of grammatical relations and $|V_{dep}|$ is about 50 in our paper.}. It is computationally expensive, even when $n=2$ . Hence, in our paper, we only include the grammatical relations for modeling $r$ which also shows the promising results in the experiment section.
		
		The proposed learning model connects two words with the dependency path in the vector space. By incorporating the dependency path into the model, the word embeddings encode more syntactic information \cite{levy-goldberg:2014:P14-2} and the distributed representations of dependency paths are explicitly learned. To be specific, this model requires that $\textbf{r}$ is close to $\textbf{w}_2 - \textbf{w}_1 $ and we minimize the following loss function for the model learning:
		\begingroup\makeatletter\def\f@size{9}\check@mathfonts
		
		$$ \sum_{(w_1,w_2,r)\in{C_1}} \sum_{r' \sim{p(r)}} max\{0,1-(\textbf{w}_2 - \textbf{w}_1)^\intercal\textbf{r} + (\textbf{w}_2 - \textbf{w}_1)^\intercal\textbf{r}'\}\eqno{(1)}$$\endgroup 
		where $C_1$ represents the set of triples extracted from the dependency trees parsed from the text corpus, $r$ is a sequence of grammatical relations that is denoted as $(g_1,g_2,...,g_n)$, $n$ is the hop number of $r$, $g_i$ is the $i$-th grammatical relation in $r$, and $p(r)$ is the marginal distribution for $r$. The given loss function encourages that the triple $(w_1,w_2,r)$ has a higher ranking score than the randomly chosen triple $(w_1,w_2,r')$. The ranking score is measured by the inner product of vector $\textbf{r} / \textbf{r}'$ and vector $\textbf{w}_2 - \textbf{w}_1$.

		We employ the recurrent neural network to learn the compositional representations for multi-hop dependency paths. The composition operation is realized with a matrix \textbf{W}:
		$$ \textbf{h}_i = f(\textbf{W}[\textbf{h}_{i-1};\textbf{g}_i]) \eqno{(2)}$$	
		\begin{equation*}
		f(x)=
		\begin{cases}$$
		1 &\mbox{if $   $ $x>1$}\\
		x &\mbox{if $ $ $-1 \leq x \leq  1$}\\
		-1 &\mbox{if $   $ $x<-1$} 
		$$ 
		\end{cases}\eqno{(3)}
		\end{equation*} 
		where $f$ is a hard hyperbolic tangent function (\emph{hTanh}), $[\textbf{a};\textbf{b}]$ is the concatenation of two vectors, $\textbf{g}_i$ is the embedding of $g_i$. We set $\textbf{h}_1=\textbf{g}_1$ and recursively perform composition operation to obtain the final representation $\textbf{r}=\textbf{h}_n$. Figure 1 (Left) illustrates the composition progress of the dependency path from \emph{service} to \emph{professional}. In this work, we train the triples whose hop numbers are less than or equal to 3. It is time-consuming to learn embeddings when we consider the triples with larger hop dependency paths.
		
		\subsection{Multi-Task Learning with Linear Context}
		We use the linear context to enhance word embeddings. The proposed method is based on the distributional hypothesis that words in similar contexts have similar meanings. Inspired by Skip-gram \cite{mikolov2013efficient}, we enhance word embeddings by maximizing the prediction accuracy of context word \emph{c} that occurs in the linear context of a target word \emph{w}. The method is shown in Figure 1 (Right). Following the previous setting in \cite{mikolov2013distributed}, we set the window size to 5. A hinge loss function, which has been proven to be effective in the cross-domain representation learning \cite{bollegala2015unsupervised}, is modified to the following function\footnote{We use dot product to calculate the similarity of words without using the softmax function that is employed in Skip-gram.}:
		$$ \sum_{(c,w)\in{C_2}}\sum_{c'\sim{p(w)}} max\{0,1-{\textbf{w}}^\intercal\textbf{c} + {\textbf{w}}^\intercal\textbf{c}'\} \eqno{(4)}$$ 
		where, $(c,w)\in C_2$ means that the word $c$ appears in the linear context of the target word $w$, and $p(w)$ is the marginal distribution for $w$. Every word has two roles, the target word and the context word of other target words. In this paper, we use different vectors to represent the two roles of a word.
		
		\subsection{Model Training}
		The negative sampling method \cite{collobert:2011b,mikolov2013distributed} is employed to train the embedding model. Specifically, for each pair $(w,c)$, we randomly select $k_W$ words $c'$ that do not appear in the linear context window of the target word $w$. These randomly chosen words are sampled based on the marginal distribution $p(w)$ and the $p(w)$ is estimated from the word frequency raised to the $\frac{3}{4}$ power \cite{mikolov2013efficient} in the corpus. We use the same method to optimize the Equation (1). The difference is, we sample path $r'$ separately according to the hop number and randomly select $k_{R_n}$ dependency paths for the $n$-hop path $r$. The selected dependency paths $r'$ have the same hop number with $r$. We experimentally find that $k_W$ = 5 is a trade-off between the training time and performance. Similarly, $k_{R_1}$, $k_{R_2}$ and $k_{R_3}$ are set to 5, 3 and 2.
		
		The model is trained by back-propagation. We use asynchronous gradient descent for parallel training. Following the strategy for updating learning rate \cite{mikolov2013efficient}, we linearly decrease it over our training instances. The initial learning rate is set to 0.001.

		\subsection{Aspect Term Extraction with Embeddings}
		We use the CRF model for aspect term extraction, and the features are constructed based on the learned embeddings. Besides the target word embedding features, we also utilize the features encoding the context information, (1) the linear context is within a fixed window of the target word in the linear sequence; (2) the dependency context is directly related with the target word in the dependency tree. These context features are used as complementary information to help label aspect terms.
		
		Given a sentence $s=(w_1,w_2,...,w_{|s|})$ and its dependency tree, we denote $C(w_i)$ as the function extracting the features for word $w_i$. In our case, $C(w_i) = [Emb(w_i),C_w(w_i),C_d(w_i)]$, where $Emb(w_i)$ represents the embedding of word $w_i$, $C_w(w_i)$ and $C_d(w_i)$ refer to the linear context embedding and the dependency context embedding respectively. These extracted embeddings are concatenated as CRF features.
		
		Based on the assumption that the label of a word mainly depends on itself and its neighboring words \cite{collobert:2011b}, the linear context is leveraged as additional information to help tag a word in NLP tasks. In our work, $C_w(w_i)$ is defined as:
		\begingroup\makeatletter\def\f@size{7}\check@mathfonts
		$$[Emb(w_{i-\lfloor len/2\rfloor}),...,Emb(w_{i-1}),Emb(w_{i+1}),...,Emb(w_{i+\lfloor len/2\rfloor})]\eqno{(5)}$$\endgroup
		where $\lfloor * \rfloor$ is a floor function, $len$ represents the window size of linear context and is set to 5 in this paper which follows~\cite{collobert:2011b}.
		
		The dependency context captures syntactic information from the dependency tree. We collect the pairs \emph{(dependency path, context word)} as the dependency context. They are directly linked with the word to be labeled. We limit the max hop number of \emph{dependency path} in pairs to 3. For example, the dependency context of target word \emph{staff} in Figure 1 is ($*\stackrel{dep}{\longrightarrow}*$, \emph{waiter}), \emph{($*\stackrel{conj}{\longrightarrow}*$, service)}, \emph{($*\stackrel{cc}{\longrightarrow}*$, and)}, ($ *\stackrel{dep}{\longrightarrow} *  \stackrel{advmod}{\longrightarrow}*$, \emph{very}) and ($ *\stackrel{dep}{\longrightarrow} *  \stackrel{amod}{\longrightarrow}*$, \emph{professional}). We sum the vectors of \emph{dependency path} and \emph{context word} as the embedding for the pair, and average the embeddings of collected pairs to derive the dependency context embedding features $C_d(w_i)$.
		
		The goal of embedding discretization is to map the real-valued embedding matrix $M_R^{d\times num}$ into the discrete-valued embedding matrix $M_D^{d\times num}$, where $d$ is the dimension of embeddings and $num$ is the size of vocabulary. Formally, the operation of embedding discretization is shown as follows:
		\begingroup\makeatletter\def\f@size{9}\check@mathfonts
		$$M_D^{ij} = \lfloor \dfrac{(M_R^{ij} - min(M_R^{i*}))\times l}{max({M_R^{i*})-min({M_R^{i*}})}}\rfloor \eqno{(6)}$$
		where $max(M_R^{i*})$ and $min({M_R^{i*}})$ are the maximum and minimum respectively in the $i$-th dimension, $l$ is the number of discrete interval for each dimension. By this transformation, we obtain CRF features for aspect term extraction, without losing much information encoded in the original embeddings.
		
		
		\section{Experiment}
		
		\subsection{Dataset and Setting}
		We conduct our experiments on the SemEval 2014 and 2015 datasets. The SemEval 2014 dataset is from two domains, laptop (D1) and restaurant (D2) and the SemEval 2015 dataset only includes the restaurant domain (D3). The details of datasets are described in Table 1, and the F1 score is used as the evaluation metric.
		
		\begin{table}[t]
			\scalebox{0.85}{
			\begin{tabular}{|l|c|c|}
				\hline
				\textbf{Dataset} & \textbf{\# Sentences} & \textbf{\# Aspects} \\
				\hline
				\hline
				D1-Train & 3045 & 2358\\
				D1-Test	 &	800 & 654\\ 
				D2-Train	& 3041 & 3693 \\
				D2-Test & 800 & 1134 \\
				D3-Train & 1315 & 1192 \\
				D3-Test & 685 & 678 \\
				\hline
			\end{tabular}
		}
			\caption{ Statistics of the SemEval 2014 and 2015 datasets.}
		\end{table}
		
		We also introduce two in-domain corpora for learning distributed representations of words and dependency paths. The corpora contain Yelp dataset\footnote{https://www.yelp.com/academic\_dataset} and Amazon dataset\footnote{https://snap.stanford.edu/data/web-Amazon.html} which are in-domain corpora for restaurant domain and laptop domain respectively. As the size of laptop reviews is small in Amazon corpus, we add the reviews of other similar products in laptop corpus. These products consist of \emph{Electronics}, \emph{Kindle}, \emph{Phone} and \emph{Video Game}. All corpora are parsed by using the Stanford corenlp\footnote{http://nlp.stanford.edu/software/corenlp.shtml}. The derived tokens are converted to lowercase, the words and dependency paths that appear less than 10 are removed. The details of unlabeled data are shown in Table 2.
		
		\begin{table}[t]
			\scalebox{0.9}{
				\begin{tabular}{|l|c|c|}
					\hline
					\textbf{Dataset} & \textbf{\# Sentences} & \textbf{\# Tokens} \\
					\hline
					\hline
					Yelp	& 14M & 310M \\
					Amazon & 13M & 280M \\
					\hline
				\end{tabular}
			}
			\caption{Statistics of unlabeled data.}
		\end{table}
		
		In order to choose $l$ and $d$, we use 80\% sentences in training data as training set, and the rest 20\% as development set. The dimensions of word and dependency path embeddings are set to 100. Larger dimensions get similar results in the development set but need more training time. $l$ is set to 15 that performs best in the development set.	
		
		We use an available CRF tool\footnote{https://crfsharp.codeplex.com/} for tagging aspect terms, and set the parameters with default values. The only exception is the regularization option. L1 regularization is chosen as it obtains better results than L2 regularization in the development set.

		\subsection{Result and Analysis}
		We compare our method with the following methods:
		
		(1) Naive: Token in test sentences is tagged as an aspect term, if it is in the dictionary that contains all the aspect terms of training sentences.
		
		(2) Baseline Feature Templates: Most of features used in the Baseline Feature Templates are adopted from the NER Feature Templates \cite{guo2014revisiting} which are described in Table 3.
		
		(3) IHS\_RD: IHS\_RD is the top system in D1 which also relies on CRF with a rich set of lexical, syntactic and statistical features. Unlike other systems in SemEval 2014 and our work, IHS\_RD trains the CRF model on the reviews of both the restaurant domain and laptop domain.
		
		(4) DLIREC: DLIREC is the top system in D2 which relies on CRF with a variety of lexicon, syntactic, semantic features derived from NLP resources. In terms of features based on the dependency tree, they use the head word, head word POS and one-hop dependency relation of the target word. Besides, different from our work, DLIREC utilizes the corpora of Yelp and Amazon reviews to derive cluster features for words.
		
		(5) EliXa: EliXa is the top system in D3 which addresses the problem using an averaged perceptron with a BIO tagging scheme. The features used in EliXa consist of n-grams, token shape (digits, lowercase, punctuation, etc.), previous prediction, n-gram prefixes and suffixes, and word clusters derived from additional data (Yelp for Brown and Clark clusters; Wikipedia for word2vec clusters).
		
		The statistical significance tests are calculated by approximate randomization, as described in \cite{Yeh:2000:MAT:992730.992783} and the results are displayed in Table 4. Compared with Baseline Feature Templates, the top systems (DLIREC, IHS\_RD and EliXa) obtain the F1 score gains of $3\%$, $2\%$ and $5\%$ in D1, D2 and D3 respectively. It indicates that high-quality feature engineering is important for aspect term extraction. In terms of embedding-based features, the target word embedding performs worse than other feature settings. Both “W+L” and “W+D” improve performances, as they capture the additional context information. By combining embedding features of target word, linear context and dependency context, we achieve comparable performance with the best systems in SemEval 2015, and outperform the best systems in SemEval 2014. It shows that, (1) with the features based on distributed representations, we can achieve state-of-the-art results in aspect term extraction; (2) the context embedding features offer complementary information to help detect aspect terms. Additionally, adding the traditional features (Baseline Feature Templates), we get better performances. It can be explained by that the traditional features capture the surface information (e.g, stem, suffix and prefix) which is not explicitly encoded in the embedding features.

		\begin{table}[t]
			\scalebox{0.85}{
				\begin{tabular}{|l|}
					\hline
					Baseline Feature Templates \\
					00:$w_{i+k}, -2\leq k \leq 2$ \\
					01:$t_{i+k}, -2\leq k \leq 2$ \\
					02:$Prefix(w_{i+k},length), -2\leq k \leq 2, 1\leq length \leq 4$ \\
					03:$Suffix(w_{i+k},length), -2\leq k \leq 2, 1\leq length \leq 4$ \\
					04:$Stem(w_{i+k}),-2\leq k \leq 2$ \\
					05:$CapitalTag(w_{i+k}),-2\leq k \leq 2$ \\
					\hline
					\hline
					Unigram Features\\
					$y_i\circ 00-05$ \\
					\hline
					Bigram Features\\
					$y_{i-1}\circ y_{i}$ \\
					\hline
				\end{tabular}
			}
			\caption{Baseline Feature Templates. $t$ is POS tag. \emph{Prefix} and \emph{Suffix} are first and last \emph{length} characters of the word. \emph{Stem} is stem of the word. \emph{CapitalTag} indicates if the first character of the word is capitalized.}
		\end{table}
		
		\begin{table}[t]
			\scalebox{0.95}{
				\begin{tabular}{|l|l|l|l|}
					
					\hline
					\textbf{Setting} & \textbf{D1} & \textbf{D2} & \textbf{D3} \\
					\hline
					\hline
					Naive & 35.64 & 47.15 & 48.08 \\
					Baseline Feature Templates & 70.72  & 81.56 & 64.32 \\
					IHS\_RD (Top system in D1) &  \underline{74.55} & 79.62 & - \\
					DLIREC (Top system in D2)&73.78 & \underline{84.01} & -\\
					EliXa (Top system in D3) & - & - & \underline{70.04} \\
					\hline
					\hline
					W & 70.34 & 79.23 & 63.53\\
					W+L & 73.72 & 83.52 & 68.27\\
					W+D & 72.56  & 81.57 & 67.13 \\
					\hline
					\hline
					W+L+D & \textbf{74.68}$\ast[5pt]$ & \textbf{84.31}$\dagger[4pt]$ & \textbf{69.12}$\dagger[4pt]$ \\
					W+L+D+B & \textbf{75.16}$\ast[5pt]$ & \textbf{84.97}$\ast[5pt]$ & \textbf{69.73}$\ast[5pt]$\\
					\hline
				\end{tabular}
			}
			\caption{Comparison of F1 scores on the SemEval 2014 and 2015 datasets. W represents the embedding features of target word. L denotes linear context embedding features. D refers to dependency context embedding features. B is Baseline Feature Templates. The marker $\ast[5pt]$ refers to p-value $<$ 0.05, the marker $\dagger[4pt]$ refers to p-value $<$ 0.01, and the compared baseline is the feature setting W.}
		\end{table}
		
		\subsection{Comparison of Different Embedding Methods}	
		We conduct experiments to evaluate our method of word and dependency path embeddings (WDEmb) with state-of-the-art embedding methods.
		
		The baseline embedding algorithms include Dependency Recurrent Neural Language Model (DRNLM)~\cite{Piotr2015}, Skip-gram, CBOW \cite{mikolov2013efficient} and Dependency-based word embeddings (DepEmb) \cite{levy-goldberg:2014:P14-2}. The DRNLM predicts the current words given the previous words, aiming at learning a probability distribution over sequences of words. Note that, besides the previous words, DRNLM also uses one-hop dependency relation as the additional features to help the model learning. Skip-gram learns word embeddings by predicting context word given the target word, while CBOW learns word embeddings by predicting current word given the context words. DepEmb learns word embeddings using one-hop dependency context. It is similar to our method to encode functional information into word embeddings. But the dependency path considered is one-hop rather than multi-hop.
		
		The implementations of baseline models are publicly released\footnote{DepEmb (https://levyomer.wordpress.com/software/)\newline Skip-gram and CBOW (https://code.google.com/p/word2vec/)  
		\newline DRNLM (https://github.com/piotrmirowski/DependencyTreeRnn) . }, and are used to train embeddings on the same corpora as our model. We set the parameters of baselines to ensure a fair comparison\footnote {The dimensions and interval numbers of baselines are tuned in the development set. The interval numbers of DRNLM, Skip-gram and CBOW are set to 10, and the interval number of DepEmb is set to 15. All dimensions of baselines are set to 100. The window sizes of Skip-gram and CBOW are set to 5, which is the same as our model of multi-task learning with linear context.}.
		As the baselines do not learn the distributed representation of multi-hop dependency path, we compare our model with them in the feature settings (W and W+L) where only word embeddings are used. The statistical significance tests are calculated by approximate randomization and the results are shown in Table 5.
		
		Among these baselines, the DepEmb learns word embeddings by dependency context, and performs best in both domains. It indicates that the syntactic information is more desirable in aspect term extraction. WDEmb outperforms DepEmb and the reasons are that, (1) compared with DepEmb that embeds a unit \emph{word} + \emph{dependency path} into a vector, WDEmb learns the word and dependency path embeddings separately, and alleviates the data sparsity problem; (2) multi-hop dependency paths are taken into consideration in WDEmb which help encode more syntactic information. DRNLM performs worst among these baselines, though it utilizes the dependency relation in their model. The reason is that the DRNLM focuses on language modeling and does not include subsequent words which are used in other embedding baselines, as the context.
		
		We also employ two methods for qualitative analysis of the learned embeddings, (1) we select the 6 most similar words (by cosine similarity) for each aspect word. Results are shown in Table 6; (2) we design some queries made up of one word and one dependency path, such as \emph{delicious + $*\stackrel{amod}{\longleftarrow}*$ } and \emph{delicious + $*\stackrel{acomp}{\longleftarrow}*\stackrel{nsubj}{\longrightarrow}*$ }, to find similar words. Results are displayed in Table 7.
		
		Table 6 shows that the similar words derived from both Skip-gram and CBOW, are less similar with the defined aspect words in syntactic function, such as, \emph{fast}, \emph{estimated} and \emph{speedy}. The reason lies in that both the Skip-gram and CBOW models learn word embeddings based on linear context and capture less functional similarity between words than the DepEmb and WDEmb~\cite{levy-goldberg:2014:P14-2}. Table 6 also shows that the similar words induced from WDEmb are more topical similar with the aspect words than DepEmb (e.g., \emph{racquetball}, \emph{everything},\emph{ smog}). The reason is that WDEmb encodes more topical information into word embeddings by multi-task learning based on linear context. Besides, we can obtain sensible words by the queries made up of \emph{word + dependency path} in Table 7. It reveals that our approach of word and dependency path embeddings is promising.
		
		\begin{table}[t]
			\scalebox{0.80}{
			\begin{tabular}{|l|l|l|l|l|l|l|}
				\hline
				\multirow{2}{*}{\textbf{Embedding}} & \multicolumn{2}{|c|}{\textbf{D1}} & \multicolumn{2}{|c|}{\textbf{D2}} & \multicolumn{2}{|c|}{\textbf{D3}}\\
				\cline{2-7}
				& \centering{W+L} & W & W+L & W & W+L & W\\
				\hline
				\hline
				DRNLM & 66.91 & 65.23 & 78.59 & 77.42 & 64.75 & 61.12 \\
				Skip-gram	& 70.52  & 67.25 & 82.20 & 78.62 & 66.98 & 62.27\\ 
				CBOW	& 69.80 & 66.61 & 81.98 & 78.27 & 67.09 & 62.34 \\
				DepEmb & 71.02 & 68.36 &  82.78 & 78.83 & 67.55 & 62.94\\
				\hline
				\hline
				WDEmb & \textbf{73.72}$\ast[5pt]$ & \textbf{70.34}$\dagger[4pt]$ & \textbf{83.52}$\dagger[4pt]$ & \textbf{79.23}$\ast[5pt]$ &\textbf{68.27}$\ast[5pt]$ & \textbf{63.53}$\ast[5pt]$\\
				\hline
			\end{tabular}
		}
			\caption{Results of different embedding methods. The marker $\ast[5pt]$ represents p-value $<$ 0.05, the marker $\dagger[4pt]$ represents p-value $<$ 0.01, and the compared baseline is DepEmb.}
		\end{table}
		
		\begin{table}[t]
			\renewcommand{\multirowsetup}{\centering}  
			\scalebox{0.65}{
				\begin{tabular}{|c|c|c|c|c|c|}
					\hline 
					& \textbf{DRNLM} &\textbf{Skip-gram} & \textbf{CBOW} & \textbf{DepEmb} & \textbf{WDEmb} \\
					\hline
					\hline
					\multirow{6}{*}{\textbf{\large food}} & place & service &  cuisine & salmonella &  sushi \\
					& meal & cuisine & service & cuisine&pizza  \\
					&  \color{blue}{way} & sushi & drinks & dimsum &meal \\
					& \color{red}{is} & \color{red}{fast} & grub & \color{blue}{racquetball} &cuisine \\
					& service &\color{red}{consistency} & \color{red}{fast} & workmanship & coffee\\
					&  drinks & drinks & vittles  & \color{blue}{everything} &  beer\\
					\hline
					\multirow{6}{*}{\textbf{\large delivery }} & wine & pickup & pickup & takeout & takeout \\
					& prices & turnaround & turnaround & turnaround & pickup \\
					& \color{blue}{grab} & carryout & shipping & towing & payment \\
					& \color{blue}{choice} & payment &  takeout & \color{blue}{smog} & shuttle \\
					& pickup & \color{red}{estimated} & \color{red}{speedy} & \color{blue}{installation} & turnaround \\
					& lunch & checkin & deliveries & concession & beverage\\
					\hline
					\multirow{6}{*}{\textbf{\large cpu}} & computer & gpu &  processor & netbook & gpu \\
					& \color{red}{266mhz} & computer & pc & hdd & processor\\
					& pc & processor & gpu & visor & core\\
					& \color{red}{works} & q6600 & \color{red}{266mhz} & \color{blue}{ibook} & motherboard \\
					& bios & \color{red}{200mhz} & cpus & mobo & computer \\
					& cpus & \color{red}{100mbps} & motherboard & bios & hdd \\
					\hline
				\end{tabular}
			}
			\caption{Aspect words and their most similar words. The words that are less similar with the aspect words in syntactic function are highlighted in red. The words that are less similar with the aspect words in topical information are highlighted in blue.}
		\end{table}
		
		\begin{table}[t]
			\renewcommand{\multirowsetup}{\centering}  
		    \scalebox{0.90}{
			\begin{tabular}{|c|c|c|}
				\hline 
				\emph{delicious} & \emph{delicious} & \emph{delicious}\\
				+&+&+\\
				\emph{$*\stackrel{amod}{\longleftarrow}*$} & \emph{$*\stackrel{acomp}{\longleftarrow}*\stackrel{nsubj}{\longrightarrow}*$}& \emph{$*\stackrel{acomp}{\longleftarrow}*\stackrel{nsubj}{\longrightarrow}*\stackrel{conj}{\longrightarrow}*$}\\
				\hline
				\hline
				cheese &  fries &  salad\\
				chips &  sauce & chicken\\
				onions &  salad & flavors\\
				potatoes &  chips & cheese\\
				spinach &  cheese & sausage\\
				rice &  flavor & apple\\
				\hline
			\end{tabular}
			}
			\caption{Queries and their most similar words. }
		\end{table}
		\begin{figure}
			\centering
			\includegraphics[width=0.95\textwidth]{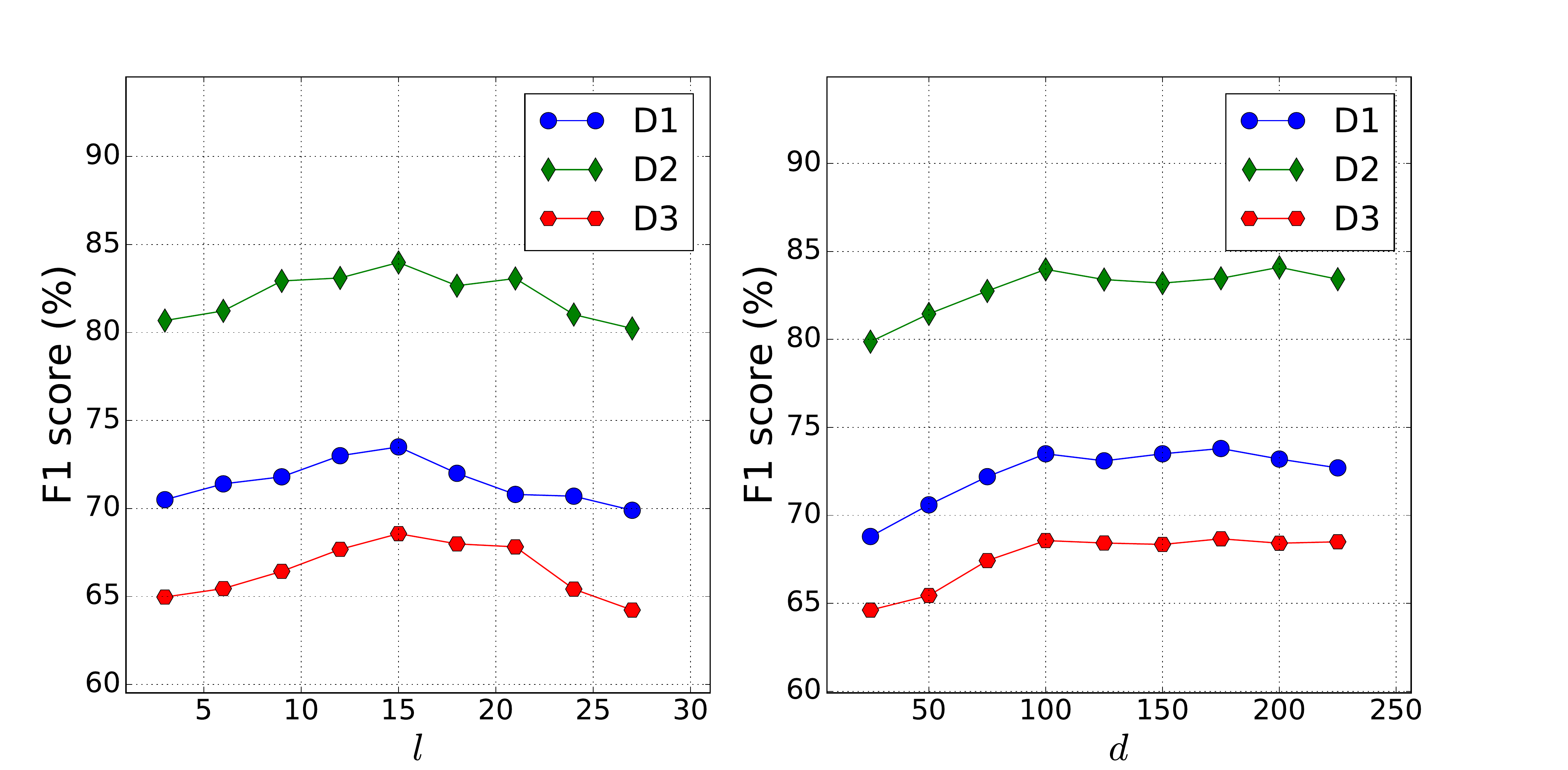}
			\caption{Left: F1 scores of our method with different $l$. Right: F1 scores of our method with different $d$.}
			\label{fig:Figure2}
		\end{figure}
		\subsection{Effect of Parameters}
		In this section, we investigate the effect of interval number $l$ and the dimension of embeddings $d$ in the feature setting of W+L+D. All the experiments are conducted on the development set.
		
		We vary $l$ from 3 to 21, increased by 3. The performances in three datasets are given in Figure 2.
		We can see that the performances increase when $l<15$, and the models achieve the best results in two domains when $l=15$. However, when $l>15$, the performances decrease.
		We also vary $d$ from 25 to 225, increased by 25. The results are shown in Figure 2. The performances increase when $d<100$. But they keep stable when $d \geq 100$. We can infer that 100 is a trade-off between the performance and training time.
				
		\section{Conclusion and Future Work}
		
		In this paper, we propose to learn the distributed representations of words and dependency path in an unsupervised way for aspect term extraction. Our method leverages the dependency path information to connect words in the embedding space. It is effective to distinguish words with similar context but different syntactic functions which are important for aspect term extraction. Furthermore, the distributed representation of multi-hop dependency paths are also derived explicitly. 
		%
		The embeddings are then discretized and used as features in the CRF model to extract aspect terms from review sentences. With only the distributed representation features, we obtain comparable results with the top systems in aspect term extraction on the benchmark dataset. Recently,~\cite{wang2015knowledge} propose to incorporate knowledge graph to represent the documents or short text~\cite{wang2015sim,wang2016knowledgekernel}. This could be interesting to combine this representation with our embedding methods to further improve the performance.
		
		


	\section{Acknowledgments}
	 This paper is partially supported by the National Natural Science Foundation of China (NSFC Grant Numbers 61272343, 61472006), the Doctoral Program of Higher Education of China (Grant No. 20130001110032) as well as the National Basic Research Program (973 Program No. 2014CB340405).
 
	\newpage
	\bibliographystyle{named}
	\bibliography{ijcai16}

\begin{thebibliography}{}

\bibitem[\protect\citeauthoryear{Bengio \bgroup \em et al.\egroup
  }{2013}]{6472238}
Y.~Bengio, A.~Courville, and P.~Vincent.
\newblock Representation learning: A review and new perspectives.
\newblock {\em PAMI}, 35(8):1798--1828, Aug 2013.

\bibitem[\protect\citeauthoryear{Bollegala \bgroup \em et al.\egroup
  }{2015}]{bollegala2015unsupervised}
Danushka Bollegala, Takanori Maehara, and Ken-ichi Kawarabayashi.
\newblock Unsupervised cross-domain word representation learning.
\newblock {\em arXiv:1505.07184}, 2015.

\bibitem[\protect\citeauthoryear{Bordes \bgroup \em et al.\egroup
  }{2011}]{bordes2011learning}
Antoine Bordes, Jason Weston, Ronan Collobert, and Yoshua Bengio.
\newblock Learning structured embeddings of knowledge bases.
\newblock In {\em Conference on Artificial Intelligence}, number
  EPFL-CONF-192344, 2011.

\bibitem[\protect\citeauthoryear{Chernyshevich}{2014}]{chernyshevich2014ihs}
Maryna Chernyshevich.
\newblock Ihs r\&d belarus: Cross-domain extraction of product features using
  conditional random fields.
\newblock {\em SemEval}, page 309, 2014.

\bibitem[\protect\citeauthoryear{Collobert \bgroup \em et al.\egroup
  }{2011}]{collobert:2011b}
R.~Collobert, J.~Weston, L.~Bottou, M.~Karlen, K.~Kavukcuoglu, and P.~Kuksa.
\newblock Natural language processing (almost) from scratch.
\newblock {\em JMLR}, 2011.

\bibitem[\protect\citeauthoryear{Guo \bgroup \em et al.\egroup
  }{2014}]{guo2014revisiting}
Jiang Guo, Wanxiang Che, Haifeng Wang, and Ting Liu.
\newblock Revisiting embedding features for simple semi-supervised learning.
\newblock In {\em EMNLP}, pages 110--120, 2014.

\bibitem[\protect\citeauthoryear{Hu and Liu}{2004}]{hu2004mining}
Minqing Hu and Bing Liu.
\newblock Mining opinion features in customer reviews.
\newblock In {\em AAAI}, volume~4, pages 755--760, 2004.

\bibitem[\protect\citeauthoryear{Jakob and
  Gurevych}{2010}]{jakob2010extracting}
Niklas Jakob and Iryna Gurevych.
\newblock Extracting opinion targets in a single-and cross-domain setting with
  conditional random fields.
\newblock In {\em EMNLP}, pages 1035--1045, 2010.

\bibitem[\protect\citeauthoryear{Levy and
  Goldberg}{2014}]{levy-goldberg:2014:P14-2}
Omer Levy and Yoav Goldberg.
\newblock Dependency-based word embeddings.
\newblock In {\em ACL}, pages 302--308, 2014.

\bibitem[\protect\citeauthoryear{Li \bgroup \em et al.\egroup
  }{2010}]{li2010structure}
Fangtao Li, Chao Han, Minlie Huang, Xiaoyan Zhu, Ying-Ju Xia, Shu Zhang, and
  Hao Yu.
\newblock Structure-aware review mining and summarization.
\newblock In {\em ACL}, pages 653--661, 2010.

\bibitem[\protect\citeauthoryear{Li \bgroup \em et al.\egroup
  }{2012}]{li2012cross}
Fangtao Li, Sinno~Jialin Pan, Ou~Jin, Qiang Yang, and Xiaoyan Zhu.
\newblock Cross-domain co-extraction of sentiment and topic lexicons.
\newblock In {\em ACL}, pages 410--419, 2012.

\bibitem[\protect\citeauthoryear{Lin \bgroup \em et al.\egroup
  }{2015}]{Lin2015Model}
Yankai Lin, Zhiyuan Liu, Huanbo Luan, Maosong Sun, Siwei Rao, and Song Liu.
\newblock Modeling relation paths for representation learning of knowledge
  bases.
\newblock In {\em EMNLP}, 2015.

\bibitem[\protect\citeauthoryear{Liu \bgroup \em et al.\egroup
  }{2015a}]{liu-joty-meng:2015:EMNLP}
Pengfei Liu, Shafiq Joty, and Helen Meng.
\newblock Fine-grained opinion mining with recurrent neural networks and word
  embeddings.
\newblock In {\em EMNLP}, pages 1433--1443, 2015.

\bibitem[\protect\citeauthoryear{Liu \bgroup \em et al.\egroup
  }{2015b}]{liu-EtAl:2015:ACL-IJCNLP}
Yang Liu, Furu Wei, Sujian Li, Heng Ji, Ming Zhou, and Houfeng Wang.
\newblock A dependency-based neural network for relation classification.
\newblock In {\em ACL}, pages 285--290, 2015.

\bibitem[\protect\citeauthoryear{Mikolov \bgroup \em et al.\egroup
  }{2010}]{mikolov2010recurrent}
Tomas Mikolov, Martin Karafi{\'a}t, Lukas Burget, Jan Cernock{\`y}, and Sanjeev
  Khudanpur.
\newblock Recurrent neural network based language model.
\newblock In {\em INTERSPEECH}, pages 1045--1048, 2010.

\bibitem[\protect\citeauthoryear{Mikolov \bgroup \em et al.\egroup
  }{2013a}]{mikolov2013efficient}
Tomas Mikolov, Kai Chen, Greg Corrado, and Jeffrey Dean.
\newblock Efficient estimation of word representations in vector space.
\newblock {\em arXiv:1301.3781}, 2013.

\bibitem[\protect\citeauthoryear{Mikolov \bgroup \em et al.\egroup
  }{2013b}]{mikolov2013distributed}
Tomas Mikolov, Ilya Sutskever, Kai Chen, Greg~S Corrado, and Jeff Dean.
\newblock Distributed representations of words and phrases and their
  compositionality.
\newblock In {\em NIPS}, pages 3111--3119, 2013.

\bibitem[\protect\citeauthoryear{Neelakantan \bgroup \em et al.\egroup
  }{2015}]{neelakantan-roth-mccallum:2015:ACL-IJCNLP}
Arvind Neelakantan, Benjamin Roth, and Andrew McCallum.
\newblock Compositional vector space models for knowledge base completion.
\newblock In {\em ACL}, pages 156--166, July 2015.

\bibitem[\protect\citeauthoryear{Piotr and Andreas}{2015}]{Piotr2015}
Mirowski Piotr and Vlachos Andreas.
\newblock Dependency recurrent neural language models for sentence completion.
\newblock {\em arXiv:1507.01193}, 2015.

\bibitem[\protect\citeauthoryear{Pontiki \bgroup \em et al.\egroup
  }{2014}]{pontiki2014semeval}
Maria Pontiki, Haris Papageorgiou, Dimitrios Galanis, Ion Androutsopoulos, John
  Pavlopoulos, and Suresh Manandhar.
\newblock Semeval-2014 task 4: Aspect based sentiment analysis.
\newblock In {\em SemEval}, pages 27--35, 2014.

\bibitem[\protect\citeauthoryear{Pontiki \bgroup \em et al.\egroup
  }{2015}]{pontiki2015semeval}
Maria Pontiki, Dimitrios Galanis, Haris Papageogiou, Suresh Manandhar, and Ion
  Androutsopoulos.
\newblock Semeval-2015 task 12: Aspect based sentiment analysis.
\newblock In {\em SemEval}, 2015.

\bibitem[\protect\citeauthoryear{Qiu \bgroup \em et al.\egroup
  }{2011}]{qiu2011opinion}
Guang Qiu, Bing Liu, Jiajun Bu, and Chun Chen.
\newblock Opinion word expansion and target extraction through double
  propagation.
\newblock {\em CL}, 37(1):9--27, 2011.

\bibitem[\protect\citeauthoryear{San~Vicente \bgroup \em et al.\egroup
  }{2015}]{sanvicente-saralegi-agerri:2015:SemEval}
I\~{n}aki San~Vicente, Xabier Saralegi, and Rodrigo Agerri.
\newblock Elixa: A modular and flexible absa platform.
\newblock In {\em SemEval}, pages 748--752, 2015.

\bibitem[\protect\citeauthoryear{Toh and Wang}{2014}]{zhiqiang2014dlirec}
Zhiqiang Toh and Wenting Wang.
\newblock Dlirec: Aspect term extraction and term polarity classification
  system.
\newblock 2014.

\bibitem[\protect\citeauthoryear{Turian \bgroup \em et al.\egroup
  }{2010}]{turian2010word}
Joseph Turian, Lev Ratinov, and Yoshua Bengio.
\newblock Word representations: a simple and general method for semi-supervised
  learning.
\newblock In {\em ACL}, pages 384--394, 2010.

\bibitem[\protect\citeauthoryear{Wang \bgroup \em et al.\egroup
  }{2015a}]{wang2015knowledge}
Chenguang Wang, Yangqiu Song, Ahmed El-Kishky, Dan Roth, Ming Zhang, and Jiawei
  Han.
\newblock Incorporating world knwledge to document clustering via heterogeneous
  information networks.
\newblock In {\em KDD}, pages 1215--1224, 2015.

\bibitem[\protect\citeauthoryear{Wang \bgroup \em et al.\egroup
  }{2015b}]{wang2015sim}
Chenguang Wang, Yangqiu Song, Haoran Li, Ming Zhang, and Jiawei Han.
\newblock Knowsim: A document similarity measure on structured heterogeneous
  information networks.
\newblock In {\em ICDM}, pages 1015--1020, 2015.

\bibitem[\protect\citeauthoryear{Wang \bgroup \em et al.\egroup
  }{2016}]{wang2016knowledgekernel}
Chenguang Wang, Yangqiu Song, Haoran Li, Ming Zhang, and Jiawei Han.
\newblock Text classification with heterogeneous information network kernels.
\newblock In {\em AAAI}, 2016.

\bibitem[\protect\citeauthoryear{Yeh}{2000}]{Yeh:2000:MAT:992730.992783}
Alexander Yeh.
\newblock More accurate tests for the statistical significance of result
  differences.
\newblock COLING, pages 947--953, 2000.

\end{thebibliography}
\end{document}